\pdfoutput=1  
\documentclass{article}

\usepackage{microtype}
\usepackage{graphicx}
\usepackage{booktabs} 

\usepackage{hyperref}

\usepackage[accepted]{mlsys2023}

\usepackage{booktabs}   
\usepackage{subcaption} 
\usepackage[utf8]{inputenc} 
\usepackage[T1]{fontenc}    
\usepackage{url}            
\usepackage{amsfonts}       
\usepackage{nicefrac}       
\usepackage{microtype}      
\usepackage{graphicx}
\usepackage{subcaption}
\usepackage{wrapfig}
\usepackage{multirow}
\usepackage{xcolor}
\usepackage{algorithmic}
\usepackage{paralist}
\usepackage{comment}
\usepackage{enumitem}
\usepackage[english]{babel}
\usepackage{epstopdf}

\usepackage{amsmath}

\newcommand{\R}{\mathbb{R}}
\newcommand{\norm}[1]{\lVert {#1} \rVert}

\newcommand{\scheme}{\textsc{Genesys}}
\newcommand{\acro}{CMA-ES}
\newcommand{\vocab}{\ensuremath{\mathbb{D}}}

\mlsystitlerunning{Synthesizing Programs with Continuous Optimization}

\begin{document}

\twocolumn[
\mlsystitle{Synthesizing Programs with Continuous Optimization}





\begin{mlsysauthorlist}
\mlsysauthor{Shantanu Mandal}{tamu}
\mlsysauthor{Todd A. Anderson}{intel}
\mlsysauthor{Javier Turek}{intel}
\mlsysauthor{Justin Gottschlich}{merly}
\mlsysauthor{Abdullah Muzahid}{tamu}
\end{mlsysauthorlist}

\mlsysaffiliation{tamu}{Department of Computer Science and Engineering, Texas A\&M University, College Station}
\mlsysaffiliation{intel}{Intel Labs}
\mlsysaffiliation{merly}{Merly AI}

\mlsyscorrespondingauthor{Shantanu Mandal}{shanto@tamu.edu}

\mlsyskeywords{Machine Learning, MLSys}

\vskip 0.3in

\begin{abstract}
Automatic software generation based on some specification is known as {\em program synthesis.} Most existing approaches
formulate program synthesis as a search problem with discrete parameters. 
In this paper, we present a {\em novel} formulation of program synthesis as a continuous optimization problem 
and use a state-of-the-art evolutionary approach, known as Covariance Matrix Adaptation Evolution Strategy to solve it.
We then propose a mapping scheme to convert the continuous formulation into actual programs.
We compare our system, called \scheme, with several recent program synthesis techniques (in both discrete and continuous domains) and show that \scheme\ synthesizes more programs within a fixed time budget than those existing schemes. For example, for programs of length 10, \scheme\ synthesizes 28\% more programs than those existing schemes within the same time budget. 

\end{abstract}

]



\printAffiliationsAndNotice{}  


\section{Introduction}
\label{sec-intro}

\emph{Program synthesis}, a subset of the field of machine programming (MP)~\cite{gottschlich:2018:mapl,
ratner:2018:sysml}, 
aims to automatically generate a software program from some specification and has the potential to revolutionize how we develop programs. 
Such systems typically produce a program as an output that satisfies some input specifications, classically in the form of input-output examples.
At a high level, most existing techniques formulate program synthesis as a search problem in the discrete domain. Typically, the search problem is solved either through formal methods~\cite{gulwani12, alur15, Bodik:2013:sttt, Cheung:2012:cikm,Heule:2016:pldi,Loncaric:2018:icse,Solar-Lezama:2006:asplos}
, machine learning approaches~\cite{deepcoder, Zohar:2018:nips, robustfill, npi, netsyn, stackgp, chen2018towards, bunel18, lgp, allgp, stackgp}
or a combination of both~\cite{feng18, nye:2019:learning}. 
Also, some prior art~\cite{difflog, stoke} has framed the problem in the continuous domain. ~\citet{difflog} formulates discrete semantics to a continuous one by annotating program rules with different weights and solve it using Markov Chain Monte Carlo technique. Recently, Neural Program Optimization (NPO)~\cite{npo} 
formulate program synthesis
in the continuous domain 
using an autoencoder~\cite{autoencoder} generated latent representations. 
Although these continuous approaches have shown promising early results for a simple problem set (e.g., programs similar to simple digital circuits such as comparator, multiplexer, etc. in case of NPO),
they fail to deliver any significant benefit 
for a complex problem set (see Section~\ref{sec-comparison}). 
Therefore, in this paper, we set out to investigate how these promising results in the continuous domain can be extended to more complex problems.



Suppose a program $P$, consisting of $l$ tokens (instructions or functions), is denoted by $P=\left<P_1, ..., P_l\right>$, where $P_i$ represents the $i$-th token for $1\le i\le l$.
$P_i$ can be any token from the set of all possible tokens, $\vocab$. 
Given some input-output examples as a specification, most existing works find a program $P$ that satisfies the specification 
by searching through different programs resulting from different combinations of various tokens. Since $P_i$ is limited to a fixed number of distinct tokens from $\vocab$, this formulation can be categorized as a discrete search problem.
Various approaches differ in the ways they prune the search space while still being able to find $P$~\cite{deepcoder,  Zohar:2018:nips, chen2021latent, hong2021latent}.
Recently, NPO proposed converting $P$ into a latent representation in the continuous domain using the encoder of an autoencoder model~\cite{npo}.
In this paper, we propose a {\em novel} formulation where $P$ can be expressed as
$P=\left<f_1(\cdot), f_2(\cdot), \dots, f_M(\cdot)\right>$. Here, each $f_i(\cdot)$, for $1\le i\le M$, is a function that takes a number of continuous parameters as inputs and maps them into some tokens in $P$. With this formulation, an error function that compares the output produced by $P$ (for a specified input) with the specified output, essentially becomes a function of continuous parameters. Therefore, the problem of program synthesis amounts to minimizing the error function but as a {\em continuous optimization} problem.


We propose to solve this continuous optimization problem using Covariance Matrix Adaptation Evolution Strategy (CMA-ES). 
Commonly used error functions for comparing program outputs, such as edit or Manhattan distance~\cite{netsyn, aip, stackgp}, are non-smooth and ill-conditioned, i.e., a small change in the input can produce a large error. Therefore, CMA-ES is perfectly suited to solve such cases.
It is a 
stochastic derivative-free algorithm for difficult (e.g., non-convex, ill-conditioned, multi-modal, rugged, noisy, etc.) optimization problems and considered as one of the most advanced 
optimization algorithms with many successful
applications~\cite{cma-es-app}. We refer to the proposed 
program synthesis framework as \scheme.

\scheme\ uses a multivariate normal distribution for generating potential solutions. 
During each generation, \scheme\ takes a number of samples from the distribution. Using the proposed {\em novel} mapping scheme, 
\scheme\ converts the samples of continuous parameters into actual programs.
These candidate programs are used to evaluate the error function. If a candidate program has no error, the target program is found. Otherwise, \scheme\ starts the next generation of evolution.
For this generation, a new mean for the distribution is calculated based on the error from the past generation. The covariance matrix, which captures dependencies among the continuous parameters, is adapted with the new mean. \scheme\ evaluates the error function with new samples and the process continues. Since CMA-ES is a local policy, 
\scheme\ can converge to a local minima. \scheme\ restarts the algorithm with new samples
to escape from the minima. 
During restarts, \scheme\ can restart CMA-ES while retaining some prior information related to the continuous parameters. \scheme\ stops when the target program is found or some maximum time limit has exceeded.
Note that NPO also uses CMA-ES to solve the optimization problem (similar to \scheme). However, NPO suffers from poor problem formulation as well as local minima due to the use of autoencoder and absence of any restart policy. 

In summary, we make the following contributions:
\begin{itemize}
    \item We investigate how program synthesis can be formulated with continuous parameters that can lead to a better synthesis algorithm.
    Towards that end, we propose a {\em novel formulation} of program synthesis as a continuous optimization problem where a program is expressed as a tuple of functions such that each function maps continuous parameters into one or more discrete tokens.

    \item We introduce a number of {\em novel mapping} schemes to convert continuous parameters into actual programs (and describe several other less performant mapping schemes). 
    We propose \scheme, a CMA-ES-based program synthesis framework that uses that mapping scheme.
    We investigate different restart policies in the context of \scheme\ to escape from local minima. 
    \item We compare \scheme\ with six prior program synthesis techniques - DeepCoder~\cite{deepcoder}, PCCoder~\cite{Zohar:2018:nips}, RobustFill~\cite{robustfill}, NetSyn~\cite{netsyn}, PushGP~\cite{stackgp}, and NPO~\cite{npo}. These techniques cover a wide spectrum of program synthesis approaches from discrete search and learning-based schemes to continuous optimization. Our results show that \scheme\ synthesizes similar or more programs within a fixed time budget than those existing schemes. Specially at higher length (e.g., 10-length), \scheme, on average, synthesizes 28.1\% more programs than those existing schemes.
\end{itemize}

\section{Background: CMA-ES}
\label{sec-background}

\acro\ was initially proposed in \cite{cmaes.1996, cmaes.2001} with a more recent version described by \cite{cmaes.hansen}. The main goal of \acro\ is to minimize an objective function $f$. 
\acro\ samples $\lambda$ points in each generation $k$ from a multivariate normal distribution $\mathcal{N}(m_k, {\sigma}^2_k.C_k)$ and adapts the parameters $C_k \in \R^{d\times d}$, $m_k \in \R^d$ and ${\sigma}_k \in \R^+$ by evaluating $f$. 
For a minimization task, $\lambda$ points are ranked by $f$ such that $f(x_1, k) \leq f(x_2, k) \leq ... \leq f(x_{\lambda}, k)$. The distribution mean is set to the weighted average $m_{k+1} = \sum^{\mu}_{i=1} \omega_i x_{i,k}$. The weights do not depend on the evaluated function value, rather on the ranking of different $\lambda$ points. Typically $\mu \leq \lambda/2$ and the weights are chosen such that $\mu_w = 1/\sum ^ \mu_{i=1}w^2_i \approx  \lambda /\ 4$~\cite{cmaes.hansen}.

Step size $\sigma_k$ is updated using cumulative step-size adaption (CSA), also known as path length control~\cite{cmaes.hansen}. The evolution path, $p_\sigma$, is updated first using Equation~\ref{eq-pq}. Then, the step size $\sigma_{k+1}$ is updated using Equation~\ref{eq-dq}.


{\scriptsize
\begin{align}
\begin{split} \label{eq-pq}
    p_\sigma = &{} (1-c_\sigma)p_\sigma + \sqrt{1-(1-c_\sigma)^2}\sqrt{\mu_w}C^{-1/2}_k  \\
              & \cdot \frac{m_{k+1}-m_k}{\sigma_k}
\end{split}\\
\begin{split} \label{eq-dq}
    \sigma_{k+1} = &{} \sigma_k \cdot e^{\frac{c_\sigma}{d_\sigma}\left(\frac{\norm{p_\sigma}}{E\norm{\mathcal{N}(0, I)}}-1\right)}.
\end{split}
\end{align}
}

Here, $c^{-1}_\sigma \approx n/3$ is the backward time horizon for the evolution path $p_\sigma$ and larger than one, $\mu_w = (\sum^\mu_{i=1} w^2_i)^{-1}$ is the variance effective selection mass and $1 \leq \mu_w \leq \mu$ holds true by the definition of $w_i$, $C^{-1/2}_k$ is the unique symmetric square root of the inverse of $C_k$, $d_\sigma$ is the damping parameter usually close to one and $E$ is the expected values of $\norm{\mathcal{N}(0, I)}$.

Finally the covariance matrix is updated, where again the respective evolution path is updated first.



{\scriptsize

\begin{align} 
\begin{split} 
    p_c = &{} (1-c_c)p_c + \boldsymbol{1}_{[0, \alpha \sqrt{n}]}(\norm {p_ \sigma})\sqrt{1-(1-c_c)^2}  \\
          & \cdot \sqrt{\mu_w}\frac{m_{k+1}-m_k}{\sigma_k}
\end{split} \\
\begin{split}
    C_{k+1} = &{} (1-c_1-c_\mu + c_s)C_k + c_1p_cp^T_c \\
              & + c_\mu \sum^\mu_{i=1}w_i \frac{x_{i:\lambda} - m_k}{\sigma_k}\left(\frac{x_{i:\lambda} - m_k}{\sigma_k}\right)^T
\end{split}
\end{align}
}

The parameters used here are as follows:
\begin{itemize}[noitemsep,topsep=0pt,parsep=0pt,partopsep=0pt, leftmargin=*]
    \item $C^{-1}_c \approx n/4$ is the backward time horizon for the evolution path $p_c$ and larger than one.
    \item $\alpha \approx 1.5$.
    \item The indicator function $\boldsymbol{1}_{[0, \alpha\sqrt{n}]}(\norm{p_\sigma})$ evaluates to $\norm{p_\sigma} \in [0, \alpha\sqrt{n}]$.
    \item $c_s = (1-\boldsymbol{1}_{[0, \alpha\sqrt{n}]}(\norm{p_\sigma})^2)c_1c_c(2-c_c)$ makes partly up for the small variance loss in case the indicator is zero.
    \item $c_1 \approx 2/n^2$ is the learning rate for the rank-one update of the covariance matrix and $c_\mu \approx \sigma_w/n^2$ is the learning rate for the rank-$\mu$ update of the covariance matrix. The covariance matrix update tends to increase the likelihood for $p_c$ and for $(x_{i:\lambda}-m_k)/\sigma_k$ to be sampled from $\mathcal{N}(0, C_{k+1})$.
\end{itemize}
CMA-ES is attractive as a powerful black box optimization technique due to the fact that it adjusts all of its parameters automatically based on the objective function and covariance matrix. Interested readers should consult \cite{cmaes.hansen} for more details about CMA-ES.

\section{Problem Statement}
\label{sec-problem}

Let $S = \{(I_j, O_j)\}^s_{j=1}$ be a set of $s$ input-output pairs, such that the output $O_j$ is obtained by executing the program $P^t$ on the input $I_j$. 
Inherently, the set $S$ of input-output examples describes the behavior of the program $P^t$.
One would like to synthesize a program $P$ that recovers the same functionality of $P^t$. However, $P^t$ is unknown, and we are provided with the set $S$ as the specification.
Based on this assumption, we define equivalency between two programs as follows:

{\bf Definition 1} (Program Equivalency).
Programs $P^a$ and $P^b$ are equivalent under the set $S = \{(I_j, O_j)\}^s_{j=1}$ of input-output examples if and only if 
$P^a(I_j) = P^b(I_j) = O_j$, for $1 \le j \le s$. We denote the equivalency by $P^a \equiv_{S} P^b$.

Definition 1 
suggests that to obtain a program equivalent to $P^t$, we need to synthesize a program that satisfies $S$.
Therefore, our goal is to find a program $P$ that is equivalent to the target program $P^t$ (which was used to generate $S$), i.e., $P \equiv_{S} P^t$.
This task is known as Inductive Program Synthesis.
Any solution to this problem requires the definition of two components. First, we need a programming language that defines the domain of valid programs. 
Second, we need a method to find $P$ from the valid program domain.


\section{\scheme\ Program Synthesis Framework}
\label{sec-main}

Here, we describe our choice of programming language, proposed formulation of program synthesis as a continuous optimization problem followed by various mapping and restart schemes 
and finally, the proposed framework, \scheme.

\begin{table}[htpb]
  \centering
  \scalebox{0.7}{
  \begin{tabular}{ll}
    \hline
    [int] & Input: \\
    \textsc{Map (+1)}  & [5, 0, -3, 1, 4] \\
    \textsc{Sort}  &  \\
    \textsc{Filter (Even)}  & Output:\\
    \textsc{Reverse}  & [6, 2, -2] \\
    \hline
  \end{tabular}}
  \caption{An example program of length 4 with an input and corresponding output.}
  \label{table:example}
\end{table}

\subsection{Domain Specific Language}
\label{sec-domain}
As \scheme's programming language, we chose a domain specific language (DSL) used in earlier work, such as NetSyn~\cite{netsyn} and DeepCoder~\cite{deepcoder}. 
This choice allows us to constrain the program space by restricting the operations used by our solution.
The only data types in the language are \emph{(i)} integers and  \emph{(ii)} lists of integers.
The DSL contains 41 functions, each taking one or two arguments and returning one output. We will refer to DSL functions as {\em Tokens} to avoid ambiguity with later terminology. We use $\vocab=\{D_i\}^{|\vocab|}_{i=1}$ to indicate the set of all tokens in the DSL, with $|\vocab|=41$. Many of the DSL tokens include operations for list manipulation. Likewise, some operations also require lambda functions. There is no explicit control flow (conditionals or looping) in the DSL. However, several of the operations are high-level functions and are implemented using such control flow structures. A full description of the DSL can be found in NetSyn~\cite{netsyn}. 
With these data types and operations, we define a program $P$ as an ordered sequence of DSL tokens i.e., $P=\left<P_i\right>^l_{i=1}$ where $l$ is the length and $P_i$ is the $i$-th token of the program. Table \ref{table:example} presents an example of a program of 4 tokens with an input and respective output. This program can be expressed as $P=\left<Map(+1), Sort, Filter(Even), Reverse\right>$.
\subsection{Program Synthesis as a Continuous Optimization Problem - A Novel Formulation}
\label{sec-cont-opt}
\begin{figure}[hptb]
\begin{center}
\includegraphics[width=0.7\columnwidth]{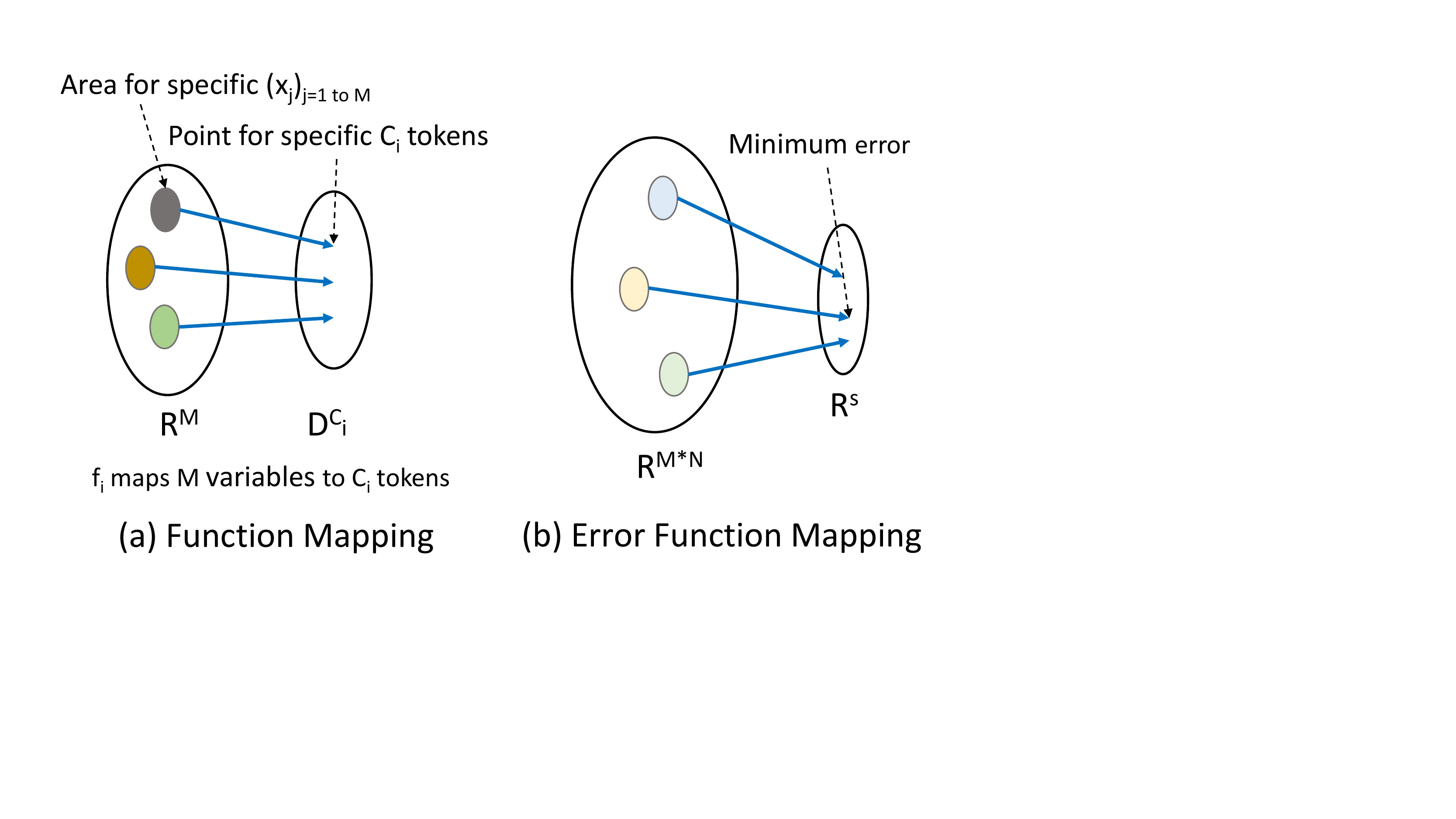}
\caption{Pictorial representation of how (a) each function maps the continuous parameters to DSL tokens in the program and (b) the error function maps the continuous parameters.    }
\label{fig_mapping}
\end{center}
\end{figure}

We propose to model a program as an $l$-tuple, 
where 
each element of the tuple is a function of continuous parameters. 
Intuitively, each function takes one or more continuous parameters and maps them to $0$ or more DSL tokens in the program.
A program $P$ can be expressed as $P=\left<f_1(x^{1}_1, ..., x^{1}_M), ..., f_N(x^{N}_1, ..., x^{N}_M)\right>$ where, $f_i(x^{i}_1, ..., x^{i}_M): \mathbb{R}^M\rightarrow \vocab^{C_i}$ for $1\le i\le N$, $0\le C_i\le l$, and $N,M>0$. Namely, each function $f_i$ maps $M$ continuous random variables into $C_i$ DSL tokens in the program. When $C_i=0$, it indicates the special case when $f_i$ does not map its parameters to any DSL token. In other words, $f_i$ becomes a NULL function.
Figure~\ref{fig_mapping}(a) shows the mapping from the continuous variables to DSL tokens. Since the length of the program $P$ is $l$,  we can infer $\Sigma_iC_i=l$.

We define an error over the given specification, $S=\{(I_j, O_j)\}$ as 
$\mathbb{E}(P, S)=\left<E(P(I_j), O_j)\right>^s_{j=1}$.
Here, $E(P(I_j), O_j)$ could be any commonly used distance function in program synthesis such as edit distance, Manhattan distance, etc.~\cite{Jurafsky2009}. With the formulation of $P$ as an $l$-tuple, $\mathbb{E}$ becomes a function mapping $\mathbb{R}^{M\times N}$ to $\mathbb{R}^{s}$. Therefore, program synthesis becomes a continuous optimization problem where the goal is to find values of $M\times N$ continuous random variables that minimize $\mathbb{E}$ (i.e., the minimum point in the $s$-dimensional space). Figure~\ref{fig_mapping}(b) depicts this formulation. 


\begin{figure}[hptb]
\begin{center}
\includegraphics[width=0.4\columnwidth]{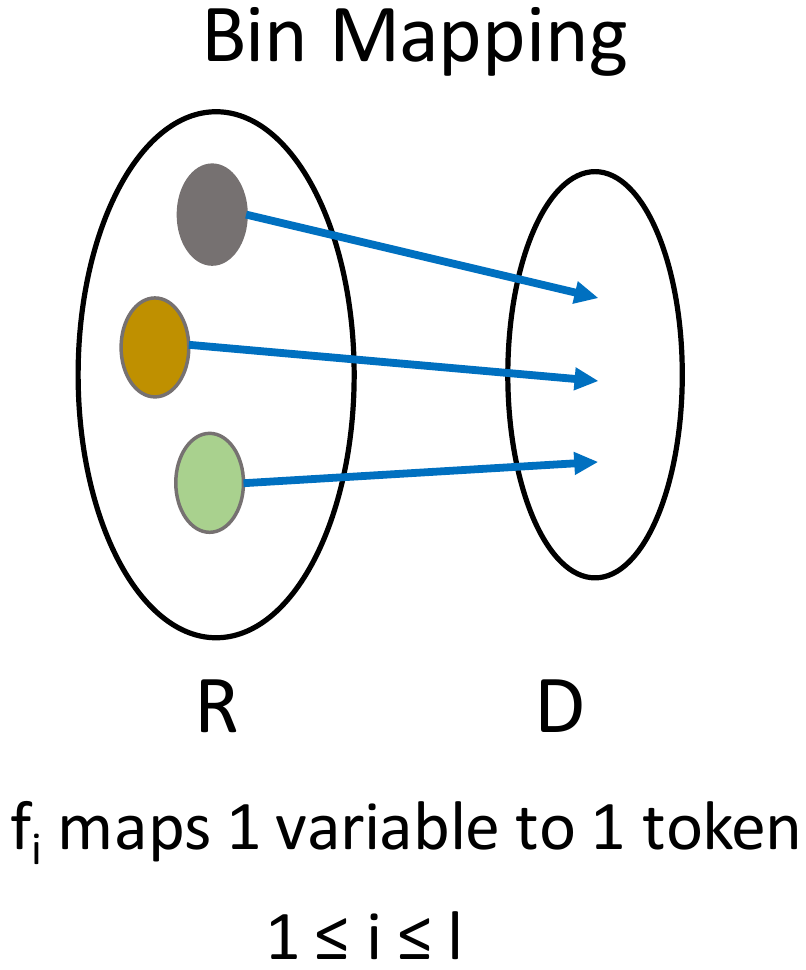}
\caption{Representation of the bin mapping scheme.}
\label{fig_bin_mapping}
\end{center}
\end{figure}

\subsection{Mapping Scheme: Bin Mapping}
\label{sec-sub-setup}
Based on the problem formulation in the prior Section, we propose a mapping scheme to construct an $l$-length program $P$.
We choose $l$ functions to represent the program.
Each function takes one continuous parameter. Thus, 
$P=\left<f_1(x^1_1), f_2(x^2_1), \dots, f_l(x^l_1)\right>$ where $f_i(x^i_1):\mathbb{R}\rightarrow\vocab$.
Each function $f_i$ must be able to map the continuous parameter to one of the DSL tokens.
This is shown in Figure~\ref{fig_bin_mapping}.
We conceptually divide the range of each variable 
into $\vert \vocab \vert$ bins. That is why, we refer to this scheme as {\em Bin Mapping}.
The size of each bin can be equal (except the bin at each end) or proportional to the probability of the corresponding token being present in the program $P$. 
Prior work~\cite{netsyn, deepcoder} showed how to infer such probability. 
When \scheme\ samples from the continuous variables to create a candidate program, it takes each real sampled value and determines into which of the bins the value falls.  The token used in that position in the candidate program is the one corresponding to the bin number into which the sampled value fell.  This sampling process occurs for each of the $l$ continuous parameters.
Thus, the definition of $f_i$ and the corresponding selection of program token is as follows: 

{\scriptsize
\begin{align*}
f_i(x^i_1) & =D_j \qquad \mathrm{if}~x^i_1\in Bin_{D_j} ~\mathrm{for}~1\le j\le |\vocab| \\
P_i & = D_j, 
\end{align*}
}

where $Bin_{D_j}$ represents the range of values corresponding to token $D_j$, and $P_i$ represents the $i$-th token of program $P$. 
This mapping scheme is robust and flexible and 
it requires the least number of variables to represent a program. This makes the optimization problem easier and helps \scheme\ to find more programs within a fixed time budget.

\subsection{Alternative Mapping Schemes}
\label{sec-alternative}
To experienced CMA-ES users, bin mapping is perhaps the most intuitive mapping but initially it was unclear whether other obviously possible mapping schemes like Multi-Group Mapping might be superior.  Thus, we did explore a number of other possible mappings before determining that bin mapping was the most effective.
Each of these other mapping schemes use one or more continuous random variables to map to $0$ or more  DSL tokens of a program. 
Table~\ref{tab_diff_mapping} summarizes 
different mapping schemes and their characteristics. 

\subsubsection{Single Group Mapping}
\label{sec-single-group}

In this mapping, $\vert \vocab \vert$ functions map to $l$-length program, $P$.
Each function represents a DSL token and takes one continuous parameter. Each function will map its parameter to at most one DSL token. Thus,
$P=\left<f_1(x^1_1), f_2(x^2_1), \dots, f_{|\vocab|}(x^{|\vocab|}_1)\right>$ where $f_i(x^i_1):\mathbb{R}\rightarrow\vocab^{0|1}$.
\scheme\ samples each continuous parameter and then constructs the program by choosing the tokens corresponding to the $l$ largest sampled values.
The token for the largest sampled value occurs first in the program.  The token corresponding to the next largest value occurs second, and so on. This is shown in Table~\ref{tab_diff_mapping}. 
A significant limitation of this approach is that each DSL token could occur at most once in a program and thus is not applicable if the length of the program, $l$ is larger than $|\vocab|$. 
The formal definition of $f_i$ and the corresponding selection of program token is as follows:

{\scriptsize
\begin{align*}
\begin{tabular}{lll}
\multirow{2}{*}{$f_i(x^i_1)=$}& $D_i$ & $if ~x^i_1 = Top_j(\{x^i_1\}^{|\vocab|}_{i=1})$ for $1\le j\le l$\\
& $ None$ & $otherwise$ \\
$P_j = $ & $D_i$ & $when ~f_i(x^i_1)=D_i$,    
\end{tabular}
\end{align*}
}



\begin{table}[h]
\centering
\resizebox{0.8\columnwidth}{!}{\begin{minipage}{\columnwidth}
\begin{tabular}{p{2cm}ccp{3.3cm}}
\toprule
Mapping Scheme & Domain & Range & Characteristics \\
\midrule
Single Group & $\mathbb{R}$ & $\vocab^{0\vert1}$ & $f_i$ maps $1$ variable to $0$ or $1$ token for $1\le i\le \vert\vocab\vert$.\\\hline
Multi-Group & $\mathbb{R}^{\vert\vocab\vert}$& $\vocab$&  $f_i$ maps $\vert\vocab\vert$ variables to $1$ token for $1\le i\le l\times\vert\vocab\vert$.\\\hline
Dynamic Multi-Group &$\mathbb{R}^{\vert\vocab\vert}$& $\vocab^{l/k}$ & $f_i$ maps $\vert\vocab\vert$ variables to $l/k$ tokens for $1\le i\le k$.\\
\bottomrule
\end{tabular}
\end{minipage} }
\caption{Characterization summary of different alternative mapping schemes.}
\label{tab_diff_mapping}
\end{table}

\subsubsection{Multi-Group Mapping}
\label{sec-multigroup}
In this mapping, \scheme\ uses $l$ functions, each taking $|\vocab|$ continuous parameters to express the program $P$.  Each position in the program is thus represented by one function. Each parameter of that function corresponds to a DSL token.
\scheme\ samples the continuous parameters of the first function and selects the largest sampled parameter. The corresponding DSL token becomes the first token in $P$. Then, \scheme\ samples the parameters of the second function and the token corresponding to the largest sampled parameter becomes the second token in $P$. This process repeats for each of the $l$ functions. 
This mapping scheme does allow the same DSL token to be used multiple times in a program but since it requires a large number of variables to represent a program, it may cause CMA-ES to take a long time to find a solution or may not find a solution. 
The formal definition of $f_i$ and the corresponding selection of program token is as follows:

{\scriptsize
\begin{align*}
f_i(x^i_1,\dots, x^i_{|\vocab|}) &= D_j \quad \mathrm{if} \,\, x^i_j = Top_1\left(\{x^i_j\}^{|\vocab|}_{j=1}\right) \\
P_i &= D_j.
\end{align*}
}

\subsubsection{Dynamic Multi-Group Mapping}
\label{sec-dynamic-multi-group}
Dynamic multi-group mapping is a hybrid of the single and multi-group mappings in which there are $k$ functions, each taking $|\vocab|$ continuous parameters and choosing $\nicefrac{l}{k}$ DSL tokens for $P$. \scheme\ randomly chooses $k$ for each program and treats $k$ as a variable that CMA-ES can evolve, thus allowing \scheme\ to evolve to find the optimal $k$.  
If $k$ is less than the program length then the continuous parameters corresponding to groups larger than $k$ will be unused.  However, since $k$ itself may increase, those continuous parameters must still be there and hence this scheme requires one more variable (i.e., $k$) than the Multi-Group Mapping.
The formal definition is as follows:  

{\scriptsize
\begin{align*}  
f_i(x^i_1,\dots, x^i_{|\vocab|}) &= \left\{D_j | ~x^i_j = Top_{\frac{l}{k}}\left(\{x^i_j\}^{|\vocab|}_{j=1}\right)\right\}\\
P_{(i-1)\frac{l}{k}+t} &= D_j \quad \mathrm{if}\,\, x^i_j = Top_t\left(\{x^i_j\}^{|\vocab|}_{j=1}\right) \quad \\ 
& \qquad\qquad\mathrm{for}\,\, 1\le t\le \frac{l}{k}.
\end{align*}
}

\subsubsection{Dynamic Bin Mapping}
\label{sec-dynamic-bin}
It is sometimes possible to find a program of a lower length (than the length of the target program) that also satisfies the given specification. This mapping scheme is the same as bin mapping but tries to take advantage of this property by including a function to choose the length, in other words the number of other bin mapping functions to turn into program statements.  The length function conceptually divides the range of $x_{l+1}^0$ into $l$ equal sized bins. When \scheme\ samples this parameter, it determines which bin the parameter's value fall into. The length corresponding to that bin is chosen as the program length.
In practice, however, we found it to be more flexible to test all subsets of a generated program of maximum length for correctness rather than evolving the length variable as this length variable is not always accurate.
The formal definition is as follows:  

{\scriptsize
\begin{align*}
f_i(x^i_1) & =D_j \qquad \mathrm{if}~x^i_1\in Bin_{D_j} ~\mathrm{and}~x^{l+1}_1\in Bin_k\\ 
& \qquad\qquad\mathrm{for}~1\le j\le |\vocab| ~\mathrm{and}~1\le k\le l ~\mathrm{and}~i\le k\\
P_i & = D_j, 
\end{align*}
}


\subsection{Restart Policy}
\label{sec-restart-policy}

\scheme\ can stop its evolutionary process when
the potential solutions are converging due to some local minima.
This is checked by determining if a change in one or multiple axes does not affect the distribution mean, or if the condition number of the covariance matrix is too high (i.e., ill-conditioned), or the evolutionary path's step size is too small to reach the solution~\cite{cmaes.hansen}. 
To escape from such situation, we investigate several restart policies where \scheme\ restarts with fresh initial parameters.
We explore all policies resulting from the combinations of 3 core restart policies: population-based (PB), mean-based (MB), and covariance matrix-based (CB).
For the PB restart, \scheme\ doubles the size of the population from its previous size. 
For the MB restart policy, \scheme\ resets the mean vector to randomly initialized values
(using the uniform random distribution) after a restart instead of keeping its current value. 
For the CB restart policy, \scheme\ re-initializes the covariance matrix to the identity matrix after a restart instead of keeping its current values. 
Additional restart policies can be constructed by any combination of these core restart policies. Some of these combinations have been proposed in earlier work such as IPOP~\cite{ipop} and BIPOP~\cite{bipop} where they re-initialize mean vector, covariance matrix, and increase population size simultaneously.



\begin{figure*}
    \begin{center}
        \includegraphics[width=0.9\textwidth]{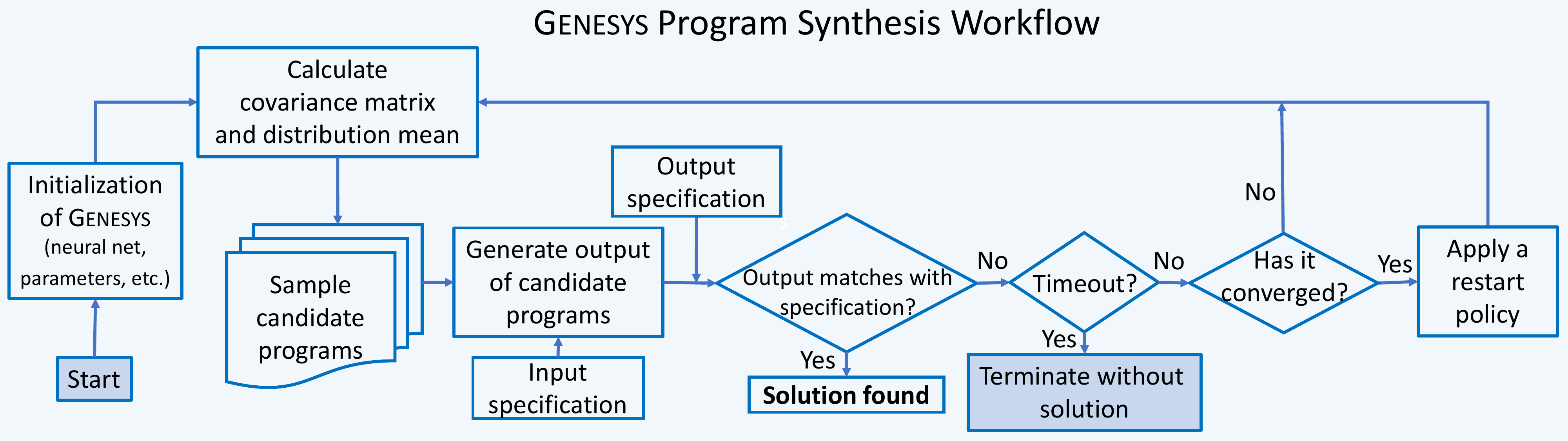} 
        \caption{Overview of \scheme.}
        \label{fig-overview}
    \end{center}
\end{figure*}

\subsection{Integration with Learning Approaches}
\label{sec-learning-desc}
Usually, CMA-ES initializes the genes by sampling continuous variables from a multivariate normal distribution and the bin mapping scheme uses a uniform size for bins to map the continuous variables.
We investigated whether \scheme\ would improve with the integration of a learning-based approach to bootstrap CMA-ES by means of better gene initialization or improve the core performance of CMA-ES by unequal bin widths based on some learned-attributes such as token probability. To do so, 
we used a neural network model similar to that of NetSyn~\cite{netsyn}. The model is a $LSTM$-based network where latent hidden state is generated as, 
$h_i = LSTM(h_{i-1}, S)$ where $S$ is the input-output example from the specification. Multiple input-output examples create different hidden states. These hidden states are aggregated together to get the token probability by passing through a $Softmax$ layer.
Details of the model and its training results are presented in Section~\ref{sec-model-character}.
\scheme\ can use the token probabilities from this model when initializing the genes at the beginning of CMA-ES.  Likewise, \scheme\ can use bin widths that are proportional to the token probabilities from this model for the bin mapping scheme.

\subsection{Putting It All Together}
\scheme\ follows the high level workflow shown in Figure~\ref{fig-overview}. \scheme\ starts by initializing a multivariate normal distribution and the population size. The continuous variables of the distribution are determined based on the bin mapping scheme in Section~\ref{sec-sub-setup}. In each generation, \scheme\ samples $\lambda$ points from the multivariate distribution and maps each sample to a program. Each of these candidate programs could potentially be the solution. Therefore, \scheme\ applies the error function $\mathbb{E}$. The error function essentially applies inputs from the given specification to a candidate program and checks if the produced output matches with the given output specification. If the outputs match, the candidate program is returned as the solution. Otherwise, \scheme\ checks if convergence has reached. If not, \scheme\ updates the mean, covariance matrix and distribution path based on the error function and starts a new generation. If, on the other hand, convergence is reached, \scheme\ applies a restart policy and repeats the whole process with new parameters. The synthesis process continues until a maximum time limit has passed or a solution is found.  




\section{Evaluation}
\label{sec-results}
\subsection{Methodology}
\label{sec-methodology}
We implemented \scheme\ in Python, starting from a base CMA-ES implementation from \url{https://github.com/srom/cma-es}.
We developed an interpreter for our DSL in Section~\ref{sec-domain} to evaluate a DSL program. 
We randomly generated a total of 600 programs from length 5 to 10, with 100 programs from each length. 
We checked those programs extensively to ensure that no program contains any token (or sequence of tokens) which does not affect the input data at all. For example, the token sequence $Map(+1), Map(-1)$ does not affect the input data at all. We performed this checking to ensure a shorter length program is unlikely to satisfy the specification of a particular length program.
We used these 600 programs as test programs that \scheme\ tries to synthesize. For each program, we used $s=5$ input-output examples as its specification.
To evaluate synthesis ability, 
we used a synthesis time limit of 3 hours (10,800 seconds) for each program. 
If any program can not be found within that time limit, we conclude that the program was ``not found''.

To run all the experiments we used SLURM batch processing system. All the experiments are run in a cluster with Intel(R) Xeon(R) CPU E5-2650 v2 @ 2.60GHz processor and 62GB of memory. 
For implementing various learning-based schemes, we use Nvidia Tesla K80 GPU with 128 GB RAM to train the neural networks.
For training any learning-based schemes, we used 420,000 randomly generated unique programs of length 5 to train the model. We checked the programs to remove duplicates. For each of the program we used 5 IO examples as the specification. For biasing bin width with token probability (Section~\ref{sec-initialize}), \scheme\ uses a neural network similar to the one used in ~\citet{netsyn} to predict the probability of various DSL functions in a target program.


\subsection{Characterization of \scheme}
\label{sec-characterization}
We first explore the characterization of 
restart policies, impact of applying learning on top of \scheme\ followed by other setups from different combinations of policies. Our main objective is to assess various settings of \scheme.

\subsubsection{Restart Policies}
\begin{figure*}[ht]
    
    \centering
        \begin{subfigure}{0.48\textwidth}
    \centering
        \includegraphics[width=0.7\textwidth]{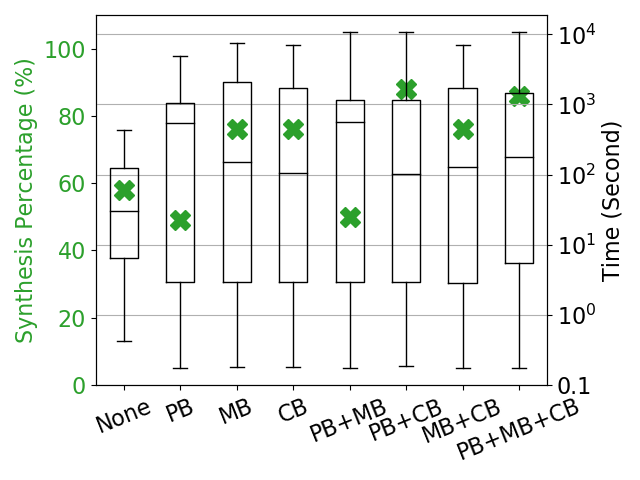}
    \caption{Restart policy}
    \label{fig-restart}
    \end{subfigure}
    ~
    \begin{subfigure}{0.48\textwidth}
    \centering
        \includegraphics[width=0.7\textwidth]{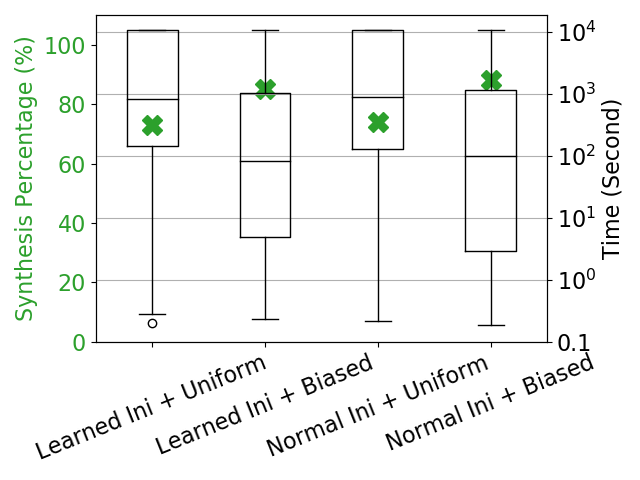}
    \caption{Impact of learning on \scheme.}
    \label{fig-initialization}
    \end{subfigure}
    
    
    \begin{subfigure}{0.48\textwidth}
    \centering
        \includegraphics[width=0.7\textwidth]{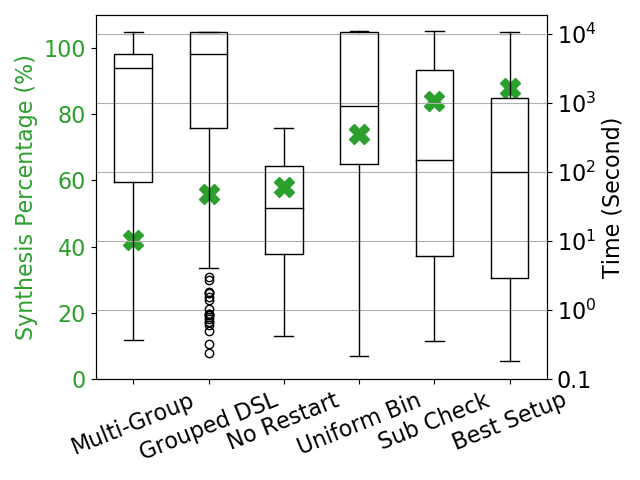}
    \caption{Different \scheme\ setup}
    \label{fig-ablation}
    \end{subfigure}
    ~
    \begin{subfigure}{0.48\textwidth}
    \centering
        \includegraphics[width=0.7\textwidth]{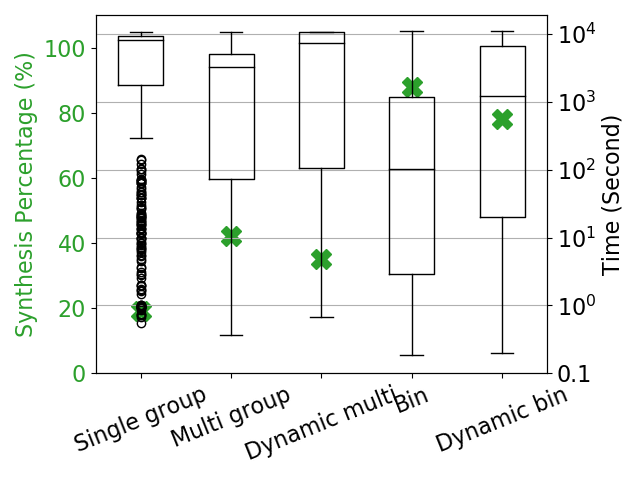}
     \caption{Mapping scheme}
     \label{fig-var}
    \end{subfigure}
    
    \caption{Synthesis percentage and time as functions of (a) restart policies (b) setup choice of learned token probability (c) other \scheme\ setups and (d) mapping schemes. Program synthesis percentages are shown (cross mark with green color) at the left $y$-axis and their synthesis times in the right $y$-axis as box plots for each of the setups. Each boxplot represents minimum, 25\% and 75\% quantiles, median and maximum synthesis times. Circles indicate outliers.}
    \label{fig:ablation_all}

\end{figure*}

The synthesis percentage and time for different combinations of the three core restart policies are presented in Figure~\ref{fig-restart}. 
Among all policies, synthesis percentage can be as low as 49\% for PB and as high as 88\% for PB+CB case. Without any restart, \scheme\ can synthesize 58\% of programs. Among the individual policies, MB and CB both synthesize $1.5\times$ more programs than that of PB. On the other hand, when we combine PB+CB, \scheme\ synthesizes 88\% programs followed by PB+MB+CB with a synthesis percentage of 86\%. Note that PB+MB+CB is essentially the BIPOP policy~\cite{bipop}. The results suggest that it is better to re-initialize the covariance matrix during restarts. This is because the covariance matrix may become ill-conditioned and lead to the same local minima.
Doubling the population is generally beneficial because more potential solutions can be checked. Two best performing policies (i.e., PB+CB and PB+MB+CB) differ by whether to re-initialize the mean vector or not. PB+CB being the best policy indicates that keeping the mean vector as it is during a restart provides some advantage 
by accumulating information about the continuous parameters.

\subsubsection{Impact of Learning}
\label{sec-initialize}

In this section we explore the effects of biasing gene initialization and bin widths in the bin mapping scheme as described in Section~\ref{sec-learning-desc}.
Figure~\ref{fig-initialization} 
shows results for different combinations of gene initialization and bin width selection.
Note from this figure that using biased bin widths is always superior in terms of synthesis percentage compared to uniform bin widths.  Also note that there is little difference between initial gene creation using a normal distribution versus a learned distribution.
Thus, we conclude that learned attributes like token probabilities are better utilized in bin width selection as opposed to the gene initialization. This is intuitive since the effect of gene initialization diminishes after a few initial iterations of \scheme. 
Thus, the default \scheme\ setup uses {\it Normal Ini + Biased} as it had a 88\% synthesis rate compared to {\it Learned Ini + Biased} which was slightly lower at 85\%.

\subsubsection{Characterization of Mapping Schemes}
Although bin mapping is the default scheme for \scheme, we investigated how other mapping schemes worked. Figure~\ref{fig-var} shows the percentage of all programs synthesized and synthesis time for different mapping schemes.
For programs of length 10, bin mapping can synthesize up to 88\% programs whereas synthesis percentage is 78\%, 19\%, 42\% and 35\% for dynamic bin, single group, multi-group and dynamic multi-group mapping respectively. In terms of synthesis time, the bin mapping scheme is at least $1.5\times$ faster than other mapping schemes. Thus, bin mapping scheme works the best both in terms of synthesis percentage and time. This is because bin mapping has the lowest number of continuous parameters (i.e., $O(l)$).
Other mapping schemes require more continuous parameters 
making the optimization domain much larger to find the solution within the given time limit.

\begin{table}[hbtp]
\centering
\scalebox{.7}{
\begin{tabular}{lccccc}
\toprule
Setup Name        & Mapping & DSL Type & Restart & Bin Type & Check \\
\midrule
Multi-Group   & Multi-group    & Regular  & Yes     & Biased      & Full            \\
Grouped DSL     & Bin  & Grouped    & Yes     & Biased      & Full            \\
No Restart    & Bin  & Regular  & No      & Biased      & Full            \\
Uniform Bin  & Bin  & Regular  & Yes     & Uniform       & Full            \\
Sub Check & Bin  & Regular  & Yes     & Biased      & Sub           \\
Best Setup    & Bin  & Regular  & Yes     & Biased      & Full            \\
\bottomrule
\end{tabular}}
\caption{Different configurations for \scheme\ framework. For restart, we used PB+CB policy.}
  \label{table:setup}
\end{table}

\subsection{Other Setups}
We experimented with additional configurations of \scheme. Table~\ref{table:setup} presents each configuration with a name. The configurations are based on 5 criteria.
{\em Mapping} and {\em Restart} indicate the mapping schemes and restart policies. {\em DSL Type} indicates whether we used our regular or grouped DSL. In grouped DSL, similar tokens are grouped into a single {\em super} token in an effort to reduce the optimization domain. {\em Bin Type} indicates whether the bin size is uniform or biased based on token probability. 
{\em Check} indicates whether \scheme\ checks for specification after a full-length program or after each token in the program (i.e., sub-program). Checking after each token allows \scheme\ to find an equivalent lower-length program at the expense of more checking operations.
Figure~\ref{fig-ablation} 
compares synthesis percentages and times for the different configurations.
The best setup synthesizes 88\% programs combining bin mapping with regular DSL, a restart policy (PB+CB), biased bins, and full program check. Token by token checking and grouped DSL did not provide as much benefit (either in terms of time or synthesis percentage) as expected due to additional checking overhead or additional selection requirements for each super token (i.e., selecting the exact token within a super token). No restart policy performs worse due to premature convergence.
Multi-group setup synthesizes the least programs because a higher number of continuous parameters lead to higher synthesis time. Therefore, many programs are not found within the given time limit. Finally, compared to the uniform bin, biased bins increase synthesis percentage by 18.9\% in the best setup. 

\subsection{Comparison of Synthesis Ability}
\label{sec-comparison}

\begin{figure*}[htpb]  
    \begin{center}
        \includegraphics[width=0.9\textwidth]{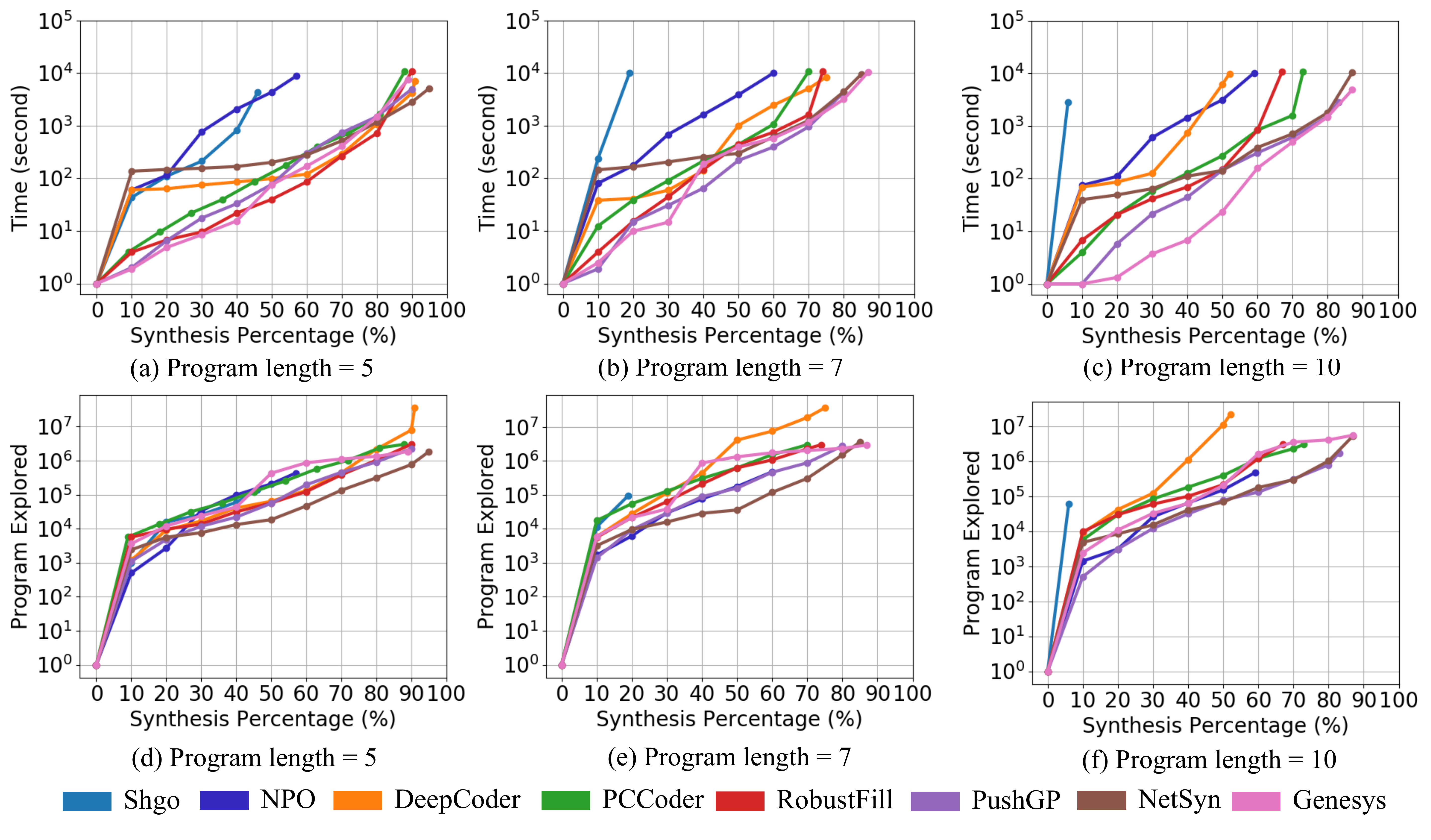}
        \caption{\scheme's synthesis ability for different program lengths with respect to Shgo, NPO, DeepCoder, PCCoder, RobustFill, PushGP, and NetSyn.}
        \label{fig:compare}
    \end{center}
    \vspace{-0.4cm}
\end{figure*}

\subsubsection{Synthesis Ability of Different Schemes}
\label{sec-syn-diff}

We compared \scheme (the best setup) with other program synthesis techniques mentioned in Table~\ref{table:schemes} .
Figure~\ref{fig:compare} shows comparative results of these methods for different program lengths. 
Synthesis time is measured in seconds as a function of the  percentage of programs synthesized within the corresponding time.
Lines terminate when an approach fails to synthesize additional programs. Among the approaches, some are learning-based while others are not. Keep in mind that \scheme\ uses a learning model to determine the bin width.
For all approaches, except Shgo and NPO, 90\% of programs can be synthesized within same amount of time for program length 5. However, as program length increases, \scheme\ synthesizes more programs in less time compared to other approaches. For example, \scheme\ synthesizes 50\% of length-10 programs in less than half of the time for other approaches. Compared to the recent techniques (DeepCoder, PCCoder, RobustFill, PushGP, and NetSyn), \scheme\ synthesizes at least 1.1\% and up to 31.3\% more programs across all lengths. 
{\em For higher length (e.g., 10-length), \scheme, on average, synthesizes 28.1\% more programs than those existing schemes}.
As expected, Shgo performs the worst since it is not used in the literature for program synthesis but is included here to show CMA-ES 
performs better than other black-box optimization techniques.
NPO is the closest work to \scheme. However, NPO has a consistently lower synthesis rate than \scheme\ for all lengths. This is due to the fact that NPO uses autoencoder generated latent variables to represent a program in the continuous domain. Even a small change in one of the latent variables can map the representation to a wildly different program leading to a large error. This characteristic makes the continuous optimization a much harder problem. On the other hand, our mapping scheme is resilient to small changes in continuous variables making the overall optimization problem easier than that of NPO. That is why, even uniform bin mapping (a scheme that does not rely on any learning model at all) can synthesize 73\% of all programs, thereby outperforming NPO, which synthesizes 58\% of all programs.
Figure ~\ref{fig:compare} (d) to (f) shows programs explored before finding the target program. Although \scheme\ searches through more programs than NetSyn, PCCoder, PushGP, and RobustFill, it searches those programs quickly. That is why, \scheme\ synthesizes more programs within fixed time. 

\begin{table}[hbtp]
    \centering
    \scalebox{.8}{
    \begin{tabular}{lp{6cm}}
    \toprule
    Scheme              & Main Idea \\
    \midrule
    Shgo                & Simplicial global optimization \\
    NPO                 & Autoencoder embedding + CMA-ES  \\
    DeepCoder           & Function probability (DNN) + DFS search      \\
    PCCoder             & Encoder decoder (DNN) + Beam search         \\
    RobustFill          & Encoder decoder (LSTM) + Beam search    \\
    PushGP              & Genetic algorithm (GA)      \\
    NetSyn              & Fitness function (LSTM) + GA      \\
    {\bf \scheme}       & Function formulation + CMA-ES + Token probability (LSTM)    \\
    \bottomrule
    \end{tabular}
    }
    \caption{Different schemes compared.}
      \label{table:schemes}
    \vspace{-0.5cm}
\end{table}

\subsubsection{Stability Comparison}
The use of autoencoder in NPO makes the problem formulation unstable.
If an optimization problem is unstable, it becomes less tolerant to random noises leading to a sharp decline in solutions. 
In order to compare stability of \scheme\ with NPO, we introduce random noises in the continuous variables. 
For example, we insert some random noises to each of the CMA-ES variable. Similarly, we introduce random noises in the latent representations of NPO. 
Table ~\ref{table:noise_tolerance} shows synthesis percentage between \scheme\ and NPO with added noise. It shows that synthesis rate for \scheme\ drops 10\% whereas, it drops 20\% for NPO on average across different length programs. This is expected as the decoder takes the latent representation from CMA-ES with noise added and produces significantly different programs. This makes it unstable. On the other hand, 
adding a little noise does not make \scheme\ wildly jump to the other bins. Rather it is kept in the same bin and \scheme\ becomes noise tolerant.

\begin{table}
    \centering
    \scalebox{0.8}{
    \begin{tabular}{ccccc}
    \toprule
    \multirow{2}{*}{Noise}     & \multirow{2}{*}{Baseline} & \multicolumn{3}{c}{Length}\\\cline{3-5}
                             &          &    5   &  7  &  10\\
    
    \midrule
    \multirow{2}{*}{Without} & \scheme & 89\%     & 87\%   & 87\% \\\cline{2-5}
                             &  NPO    & 57\%     & 60\%   & 59\%  \\\hline
                                   
    \multirow{2}{*}{With}   & \scheme  & 79\%     & 79\%   & 78\%  \\\cline{2-5}
                            &  NPO     & 51\%     & 45\%   & 42\%    \\
    \bottomrule
    \hline
    \end{tabular}
    }
    \caption{Stability comparison between \scheme\ and NPO in terms of synthesis percentage.}
      \label{table:noise_tolerance}
     \vspace{-0.7cm}
\end{table}

\section{Related Work}
\label{sec-rel}
Machine programming as known as program synthesis has been widely studied with various applications. \emph{Formal program synthesis} uses formal and rule based methods to generate programs\cite{mana75, wang2018relational}. Formal program synthesis usually guarantees some program properties by evaluating a generated program's semantics against a corresponding specification~\cite{gulwani12, alur15}. In ~\cite{feng18}, authors use a SMT based solver to find different constraints in the program and learn useful lemma that helps to prune away large parts of the search space to synthesize programs faster. However, these techniques can often be limited by exponentially increasing computational overhead that grows with the program's instruction size~\cite{Heule:2016:pldi, Bodik:2013:sttt}. 

Another way to formal methods for MP is to use machine learning (ML). Machine learning differs from traditional formal program synthesis in that it generally does not provide correctness guarantees. Instead, ML-driven MP approaches are usually only \emph{probabilistically} correct, i.e., their results are derived from sample data relying on statistical significance~\cite{Murphy:2012:mitpress}. Such ML approaches tend to explore software program generation using an objective function. Objective functions are used to guide an ML system's exploration of a problem space to find a solution. Other deep learning based program synthesizer ~\cite{bunel18, parisotto2016neuro, chen2020program} also tried different approaches such as reinforcement learning to solve the problem. Most of these works focus on synthesizing programs in domain specific languages, such as FlashFill~\cite{robustfill, kalyan2018neural} for string transformation problem, simulated robot navigation, such as Karel~\cite{shin2018improving, chen2018execution} or LIST manipulation~\cite{deepcoder, netsyn}
work.


Among the ML-based MP, in Deepcoder\cite{deepcoder}, the authors train a neural network with input-output examples to predict the probabilities of the functions that are most likely to be used in a program. Raychev et al.~\cite{raychev14} take a different approach and use an n-gram model 
to predict the functions that are most likely to complete a partially constructed program.
Robustfill~\cite{robustfill} encodes input-output examples using a series of 
recurrent neural networks (RNN), and generates the the program using another RNN one token at a time.
Bunel et al.~\cite{bunel18} explore a unique approach that combines reinforcement learning (RL) with a supervised model to find semantically correct programs. These are only a few of the works in the MP space using neural networks~\cite{netsyn, dilling-multi_spec, npi, cai17, Chen18}.

Significant research has been done in the field of genetic programming~\cite{stackgp, lgp, allgp} too, where the goal is to find a solution in the form of a complete or partial program for a given specification. Prior work in this field has tended to focus on either the representation of programs or operators during the evolution process. 
Real et al.~\cite{real18} recently demonstrated that genetic algorithms can generate accurate image classifiers. Their approach produced a state-of-the-art classifier for CIFAR-10~\cite{cifar10} and ImageNet~\cite{imagenet} datasets. Moreover, genetic algorithms have been exploited to successfully automate the neural architecture optimization process~\cite{salimans17, such17, liu17, santientlabs, real2020automlzero}.

However, all these works formulate program synthesis as a search problem of discrete parameters. On the other hand, \scheme\ tries to formulate program synthesis as a continuous optimization problem and uses a well established derivative free method, CMA-ES, to solve it. 
Previously, \acro\ was used in machine learning \cite{mueller1999a}, aerospace engineering \cite{Lutz1997}, mechanical engineering \cite{Sonoda2004}, health \cite{Winter2002} and various other engineering fields \cite{cma-es-app} but not in the field of synthesizing complex programs.

\section{Conclusion}
\label{sec-conc}
In this paper, we 
presented a novel formulation of program synthesis as a continuous optimization problem. Essentially, it expresses a program with a tuple of functions, each taking a number of continuous parameters and mapping them to zero or more DSL tokens.
We showed that 
a state-of-the-art evolutionary approach, CMA-ES, can be used to solve the optimization problem very efficiently. Our proposed framework, \scheme, consists of 
a mapping scheme to convert the continuous formulation into actual programs and restart policies for the evolutionary strategy to escape local minima. 
We compared the proposed framework with several recent program synthesis techniques and showed that \scheme\ synthesizes, on average, 28.1\% more programs within a fixed time budget at higher lengths. We envision \scheme\ to be a stepping stone towards a completely new domain of research in program synthesis.


\bibliography{main.bib}
\bibliographystyle{mlsys2023}


\newpage
\clearpage
\appendix
\section{Overview and Methodology of Previous Schemes}
\label{sec-overview-prior}

Table~\ref{table:schemes} outlines an overview of each of the schemes we reproduced.

\textbf{Shgo:} Shgo ~\cite{shgo} is simplicial homology global optimization technique. We used Shgo as a blackbox continuous optimization technique to justify our choice of CMA-ES in \scheme. We used Shgo implementation from \url{https://docs.scipy.org/doc/scipy/reference/generated/scipy.optimize.shgo.html}.

\textbf{NPO:} In NPO ~\cite{npo}, an autoencoder based neural network model is used to encode and decode programs to and from latent continuous representations.
In one end the encoder takes program and 
and creates a vector representation of the program. CMA-ES works with such vectors to find the optimal vector. That vector is passed through a decoder to produce the target program. We used stack based LSTM for the encoder and decoder. The encoder takes the program and tries to produce the same programs on the decoder part. We used 420K programs to train the autoencoder. The model took $\sim$12 hours to train.

\textbf{DeepCoder:} In Deepcoder~\cite{deepcoder}, a feed forward neural network model is used to predict function probability for a given specification. The model passes IO examples through an embedding layer 
and produces a 20 length long vector. Then it passes through multiple layer of feed forward block. All the IO examples are passed through their individual block and aggregated together with a pooling layer. A softmax function is used to predict the probability distribution of the functions. We used all the 420K random programs to train the model (Section~\ref{sec-methodology}). It took $\sim$2 days for the model to converge.

\textbf{PCCoder:} PCCoder~\cite{Zohar:2018:nips} is an encoder-decoder based neural network model. It takes the specification and creates state embedding. Those state embedding is passed through the dense block of neural networks (NN). For multiple IO examples, it passes through different NN blocks and max pooling is used to aggregate the result. Then it uses softmax to get the probability distribution. Like DeepCoder, we used all 420K programs and trained for $\sim$2 days.

\textbf{RobustFill:} RobustFill~\cite{robustfill} is a seq-to-seq neural network model that uses a encoder-decoder architecture where the encoder takes the specification and the decoder produce the program. It generates the probability distribution of different functions and a beam search is used to search for the program. The hidden representation $h_t$ is an LSTM hidden unit given by $E_t = Attention(h_{t-1}, e(X))$, $h_t = LSTM(h_{t-1}, E_t)$. Here $e(X)$ is the sequence of hidden states after processing the specification with an LSTM encoder. For multiple IO examples, all of them are passed with different network and max pooling is used to aggregate the output. Then they are passed through a softmax layer to get the probability distribution. We used all 420K programs and trained the model for $\sim$2 days to converge. 

\textbf{PushGP:} PushGP is a genetic programming based approach~\cite{stackgp}. It works in a stack based evaluator. Each of the data type has it's own stack. And recent result is always on top of the stack. Push instruction acts by pushing and popping various elements on and off the stacks. The program interpreter maintain it's own exec stack that maintain the control flow of the program. Our PushGP implementation follows the technique mentioned in \cite{stackgp}.

\textbf{NetSyn:} NetSyn \cite{netsyn} uses genetic algorithm with a LSTM based recurrent neural network as fitness function. Fitness function give some score based on the genes from gene pool about how good or bad those genes are. Based on the score, genetic mutation happens and new genes are created for next generation. The LSTM model takes IO examples and program traces as input and produce fitness score as output. It took $\sim$3 days to train the model. This LSTM model is used in one of the schemes (i.e., bin mapping with biasing (Section~\ref{sec-sub-setup})) of \scheme.

In summary, we used the same training set to train each of the prior schemes. We trained the models until they converged. Thus, we gave the best effort for a faithful reproduction of prior schemes.

\section{Others}

\begin{figure}[htpb]
    \centering
    
    \begin{subfigure}{1.0\columnwidth}
        \includegraphics[width=\textwidth]{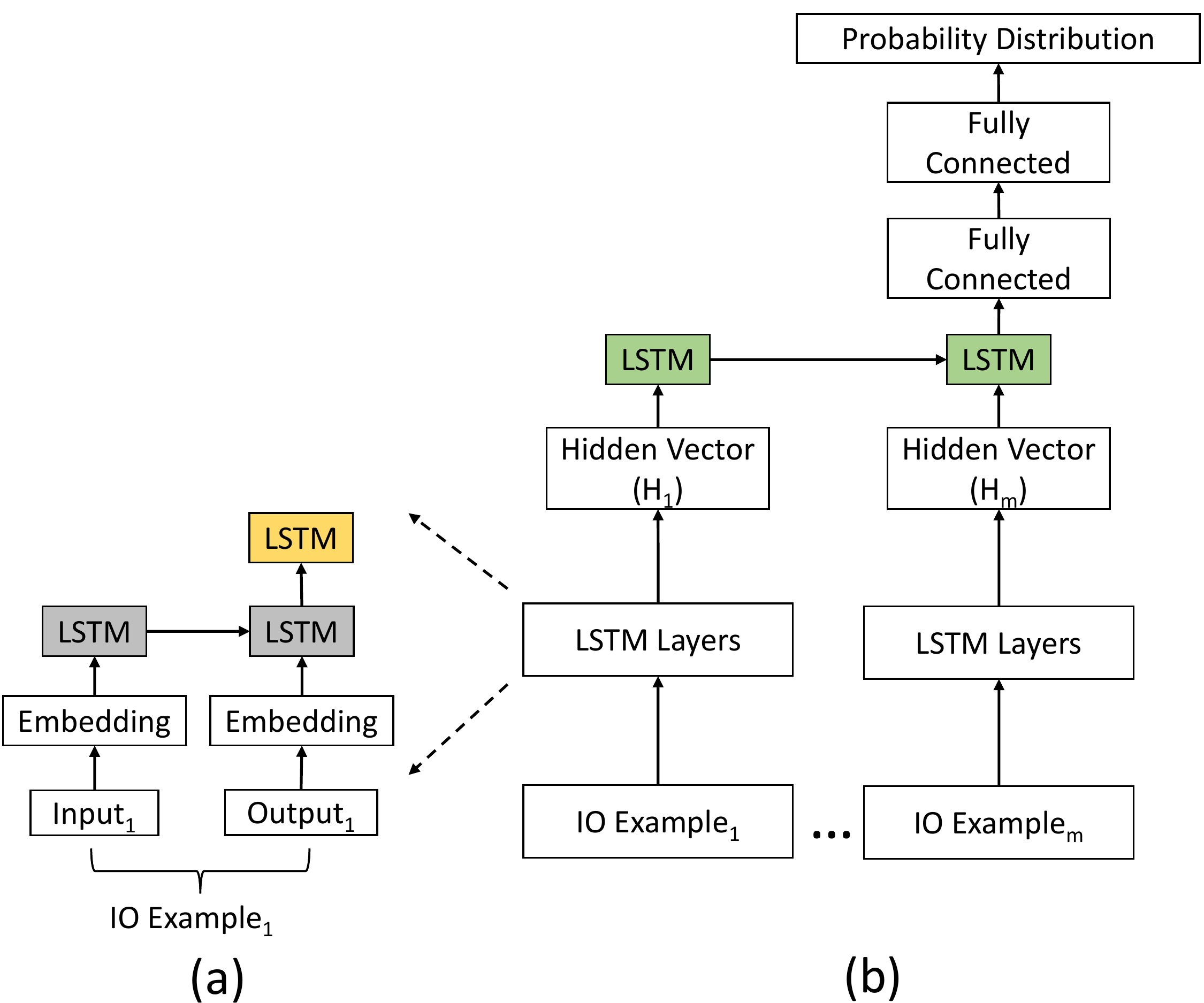}
    \end{subfigure}

    \caption{Neural network design for (a) single and (b) multiple IO examples. In each figure, layers of LSTM encoders are used to combine multiple inputs into hidden vectors for the next layer. Token probability distribution is produced by the fully connected layer.}
    \label{fig:model}
\end{figure}

\begin{figure}[htpb]
    \begin{center}
    \includegraphics[width=1.0\columnwidth]{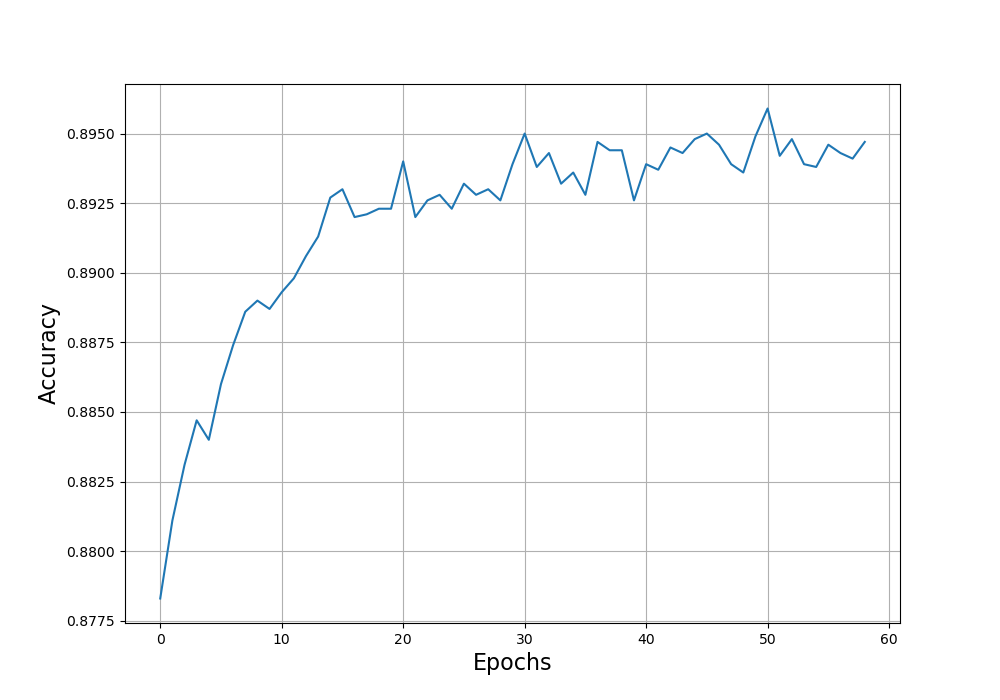}
    \vspace{-0.2cm}
    \caption{Neural network training accuracy}
    \label{fig-model-character-training-acc}
    \end{center}
\vspace{-0.4cm}
\end{figure}

\subsection{Learning Model Characterization}
\label{sec-model-character}

Figure~\ref{fig:model} shows the neural network model that we use to learn function probability. It has two components mainly. Figure ~\ref{fig:model}(a) shows how to feed single IO example to the network. Input and output is a list of integer values. They go through embedding layer and each element in the list is represented by 20 length long latent representation. After that, embedded input and output is fed through LSTM layer. We use 5 IO examples in the specification. Each of the IO create their own hidden vector from their LSTM block. Those vectors got through a different LSTM block and some fully connected layer before giving the token probability distribution for the given specification. Figure ~\ref{fig-model-character-training-acc} shows the training accuracy of the model across epochs. The model is trained until it converges. We use same number of training data as other baselines described in Section~\ref{sec-overview-prior}

\subsection{DSL grouping}

Apart from the regular DSL we tried one different DSL grouping approach. In our DSL we have two type of functions, regular function and higher order function. For example, Map is a higher order function that can have multiple regular functions like Map(+1), Map(*2) etc. In DSL group mechanism we group such higher order function together and while constructing program we tried all possible combination of functions from the same group. Although this approach might increase the synthesis time, the intuition was synthesis rate might also be increased.

\begin{figure}[htpb]
\begin{center}
\includegraphics[width=1.0\columnwidth]{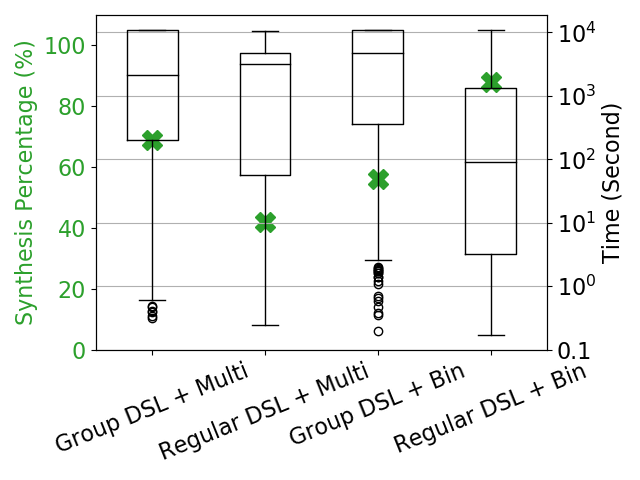}
\vspace{-0.2cm}
\caption{Choice of DSL with Multi group and Bin mapping}
\label{fig-dsl_group}
\end{center}
\vspace{-0.4cm}
\end{figure}

Figure ~\ref{fig-dsl_group} shows result for two different DSL setup that we tried. For multi group mapping, synthesis rate is higher for group DSL compared to regular DSL as 69\% vs 42\%. But when binning mapping is used we get as high as 88\% synthesis rate with regular DSL compare to 56\% for Group DSL. Synthesis time also follows the same pattern.

\subsection{Restart Number}

On average \scheme\ restarts 5-15 times for different restart policies except {\it PB} based. In {\it PB} policy, only the population get doubled up but the mean vector and covariance-matrix remain same. After reaching some local minima, despite the population increase it always stuck in the same minima for every restart as mean or covariance doesn't change. Thus, in {\it PB} policy once \scheme\ reaches some local minima it restarts in every generation and needs to stop as the increased population make the system out of memory.

\subsection{Characterizing CMA-ES Internals}
\label{sec-internals}

CMA-ES starts by initializing five major components which are called state variables. Three of them are related to covariance matrix, one is for mean vector and the last one is step size, called sigma. A sample size is pre-determined to set the number of sampled out genes from the state variables. Now, at the start of each generation, CMA-ES samples out genes. To do that, it first multiplies the eigenvectors and eigenvalues with the sampled out values from normal distribution. Mean vector is added after multiplying with sigma. After getting the fitness score for each of the genes, these state variables updated based on the weighted fitness score for the next generation.

\begin{figure*}[htpb]
    \centering
    \begin{subfigure}{0.3\textwidth}
        \includegraphics[width=\textwidth]{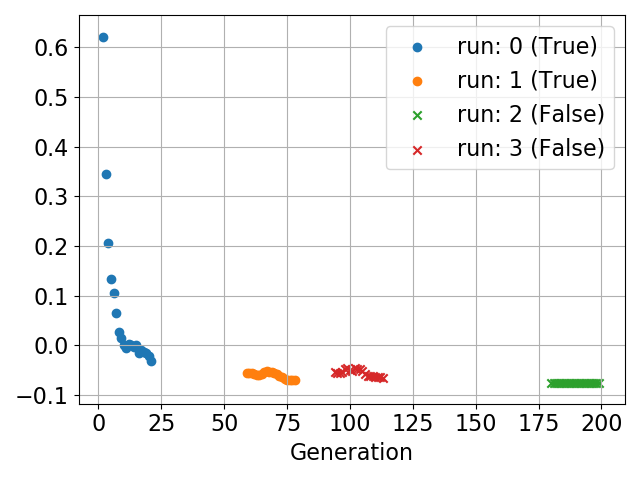}
     \vspace{-0.6cm}
     \caption{Covariance matrix}
    \end{subfigure}
    ~
    \begin{subfigure}{0.3\textwidth}
        \includegraphics[width=\textwidth]{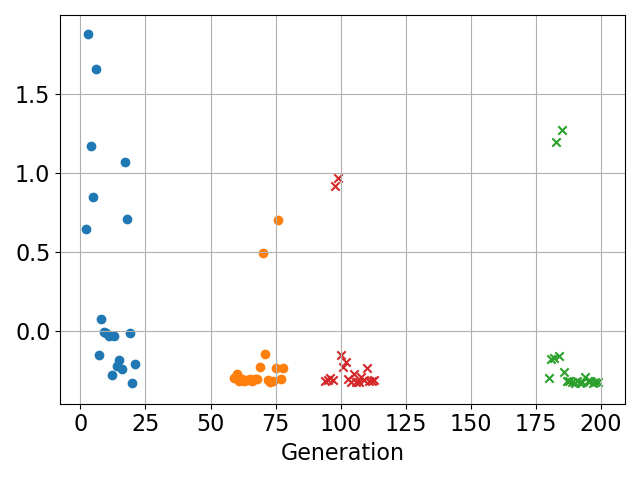}
    \vspace{-0.6cm}
    \caption{Eigenvectors}
    
    \end{subfigure}
    ~
    \begin{subfigure}{0.3\textwidth}
        \includegraphics[width=\textwidth]{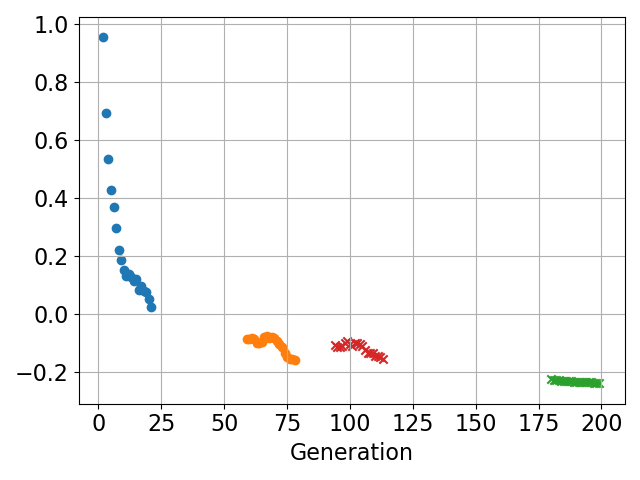}
     \vspace{-0.6cm}
     \caption{Eigenvalues}
    \end{subfigure}
    
    \begin{subfigure}{0.3\textwidth}
        \includegraphics[width=\textwidth]{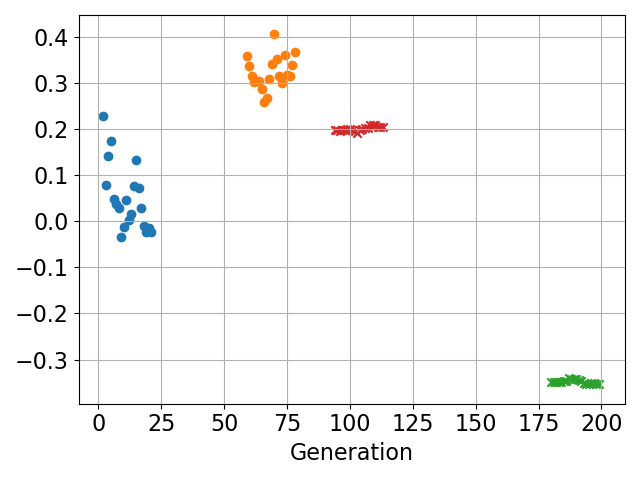}
    \vspace{-0.6cm}
    \caption{Mean}
    
    \end{subfigure}
    ~
    \begin{subfigure}{0.3\textwidth}
        \includegraphics[width=\textwidth]{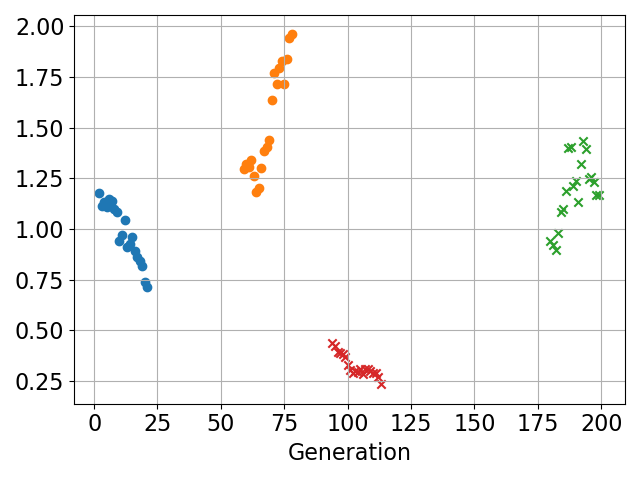}
     \vspace{-0.6cm}
     \caption{Step size}
    \end{subfigure}
    
    \caption{Here we show 5 of the major internal components of \acro. They are 3 squared matrices; Covariance matrix (C), Eigenvectors (B), Eigenvalues (D), mean and step size (sigma). We used PCA to convert the squared matrix into scalar value for plotting purpose. We choose two program cases, found case (True) and not found case (False), two from each; to show their behavior while being synthesized in \acro. We took last N generation values and plot them against corresponding generation number.}
    \label{fig:internal}
    \vspace{-0.4cm}
\end{figure*}

Figure~\ref{fig:internal} shows CMA-ES characteristics for these five major components. We choose one program to synthesize and run with \scheme\ with different random seeds. Then we took 2 runs from {\em found} and {\em not found} case to see how CMA-ES behaves internally. As covariance matrix (C), eigenvectors (B) and eigenvalues (D) are all squared matrix, for 2d visualization we convert them into scalar value using principal component analysis (PCA)~\cite{li:2014:pca}. We do the same for the mean vector. Mean vector dimension is same as the number of variables in CMA-ES. Step size is a scalar value. We run \scheme\ without any restart policy and took last N generations values for visualization purpose. Figure~\ref{fig:internal} shows that for the {\em not found} cases, 
different CMA-ES component values become stagnant and hence, \scheme\ is not able to find the solution. We notice that the mean vector is the most affected one by the stagnant values in the {\em not found} cases. The variance of the mean vector and step size is high for the {\em found} cases over generations. However, for the {\em not found} cases, it reduces significantly. 
For both cases, covariance matrix converges guided by the fitness score. Due to the ill fitness scores, it may stuck in a local minima where the solution is not present. This observation motivated us to use the restart mechanism and find the best restart policy for \scheme.

\begin{table}[hbtp]
\begin{tabular}{clc}
\hline
\multicolumn{1}{l}{Program Length} & Baseline   & \multicolumn{1}{l}{Synthesis Percentage (\%)} \\
\hline
5                                  & Shgo       & 46                                            \\
5                                  & NPO        & 57                                            \\
5                                  & DeepCoder  & 91                                            \\
5                                  & PCCoder    & 88                                            \\
5                                  & RobustFill & 90                                            \\
5                                  & PushGP     & 90                                            \\
5                                  & NetSyn     & 95                                            \\
5                                  & GeneSys    & 89                                            \\
\hline
7                                  & Shgo       & 19                                            \\
7                                  & NPO        & 60                                            \\
7                                  & DeepCoder  & 75                                            \\
7                                  & PCCoder    & 70                                            \\
7                                  & RobustFill & 74                                            \\
7                                  & PushGP     & 80                                            \\
7                                  & NetSyn     & 85                                            \\
7                                  & GeneSys    & 87                                            \\
\hline
10                                 & Shgo       & 6                                             \\
10                                 & NPO        & 59                                            \\
10                                 & DeepCoder  & 52                                            \\
10                                 & PCCoder    & 73                                            \\
10                                 & RobustFill & 67                                            \\
10                                 & PushGP     & 83                                            \\
10                                 & NetSyn     & 87                                            \\
10                                 & GeneSys    & 87                                            \\                                      
\hline
\end{tabular}
\caption{Synthesis percentage with different baselines across multiple program length}
    \label{table:detail-syn-rate}
\end{table}


\end{document}